\documentclass[conference]{IEEEtran}
\IEEEoverridecommandlockouts
\usepackage{cite}
\usepackage[colorlinks,urlcolor=blue,linkcolor=red,anchorcolor=red,citecolor=green]{hyperref}
\usepackage{amsmath,amssymb,amsfonts}
\usepackage{algorithmic}
\usepackage{algorithm}
\usepackage{graphicx}
\usepackage{multicol}
\usepackage{multirow}
\usepackage{textcomp}
\usepackage{xcolor}
\usepackage{float}
\usepackage{booktabs}
\usepackage{marvosym}
\usepackage{stfloats}
\usepackage{bbm}
\usepackage{url}
\usepackage{siunitx}
\usepackage{svg}
\usepackage{caption}
\usepackage{float}
\def\BibTeX{{\rm B\kern-.05em{\sc i\kern-.025em b}\kern-.08em
    T\kern-.1667em\lower.7ex\hbox{E}\kern-.125emX}}

\newcommand{\spheading}[2][4em]{\rotatebox{90}{\parbox{#1}{\raggedright #2}}}
\usepackage{marvosym}
\usepackage{color}

\makeatletter

\def\ps@IEEEtitlepagestyle{%
  \def\@oddfoot{\mycopyrightnotice}%
  \def\@evenfoot{}%
}
\def\mycopyrightnotice{%
  {\footnotesize 979-8-3503-3748-8/23/\$31.00~\copyright~2023 IEEE\hfill} 
  \gdef\mycopyrightnotice{}
}

\begin{document}
\title{MGCT: Mutual-Guided Cross-Modality Transformer for Survival Outcome Prediction using Integrative Histopathology-Genomic Features\\}
\author{Paper ID: B280}

\DeclareRobustCommand*{\IEEEauthorrefmark}[1]{%
  \raisebox{0pt}[0pt][0pt]{\textsuperscript{\footnotesize #1}}%
}

\author{
    \IEEEauthorblockN{
        Mingxin Liu\IEEEauthorrefmark{1}, 
        Yunzan Liu\IEEEauthorrefmark{1},
        Hui Cui\IEEEauthorrefmark{2},
        Chunquan Li\IEEEauthorrefmark{3,4,5,\Letter}  \thanks{\textsuperscript{\Letter} Corresponding authors: Chunquan Li and Jiquan Ma.} and 
        Jiquan Ma\IEEEauthorrefmark{1,\Letter} \thanks{Email: lcqbio@163.com, majiquan@hlju.edu.cn}
    }
    \IEEEauthorblockA{
        \IEEEauthorrefmark{1} Department of Computer Science and Technology, Heilongjiang University, Harbin, China\\
        \IEEEauthorrefmark{2} Department of Computer Science and Information Technology, La Trobe University, Melbourne, Australia \\
        \IEEEauthorrefmark{3} The First Affiliated Hospital, Institute of Cardiovascular Disease, Hengyang Medical School, \\ University of South China, Hengyang, China \\
        \IEEEauthorrefmark{4} The First Affiliated Hospital, Cardiovascular Lab of Big Data and Imaging Artificial Intelligence, Hengyang Medical School, \\ University of South China, Hengyang, China \\
        \IEEEauthorrefmark{5} Department of Biochemistry and Molecular Biology, School of Basic Medical Sciences, Hengyang Medical School, \\ University of South China, Hengyang, China \\
    }
}

\maketitle

\begin{abstract}
The rapidly emerging field of deep learning-based computational pathology has shown promising results in utilizing whole slide images (WSIs) to objectively prognosticate cancer patients. However, most prognostic methods are currently limited to either histopathology or genomics alone, which inevitably reduces their potential to accurately predict patient prognosis. Whereas integrating WSIs and genomic features presents three main challenges: (1) the enormous heterogeneity of gigapixel WSIs which can reach sizes as large as 150,000$\times$150,000 pixels; (2) the absence of a spatially corresponding relationship between histopathology images and genomic molecular data; and (3) the existing early, late, and intermediate multimodal feature fusion strategies struggle to capture the explicit interactions between WSIs and genomics. To ameliorate these issues, we propose the \textbf{M}utual-\textbf{G}uided \textbf{C}ross-Modality \textbf{T}ransformer (MGCT), a weakly-supervised, attention-based multimodal learning framework that can combine histology features and genomic features to model the genotype-phenotype interactions within the tumor microenvironment. To validate the effectiveness of MGCT, we conduct experiments using nearly 3,600 gigapixel WSIs across five different cancer types sourced from The Cancer Genome Atlas (TCGA). Extensive experimental results consistently emphasize that MGCT outperforms the state-of-the-art (SOTA) methods.
\end{abstract}


\section{Introduction}
\begin{figure}[t]
    \centering
    \centerline{\includegraphics[width=0.5\textwidth,height=0.5\textwidth]{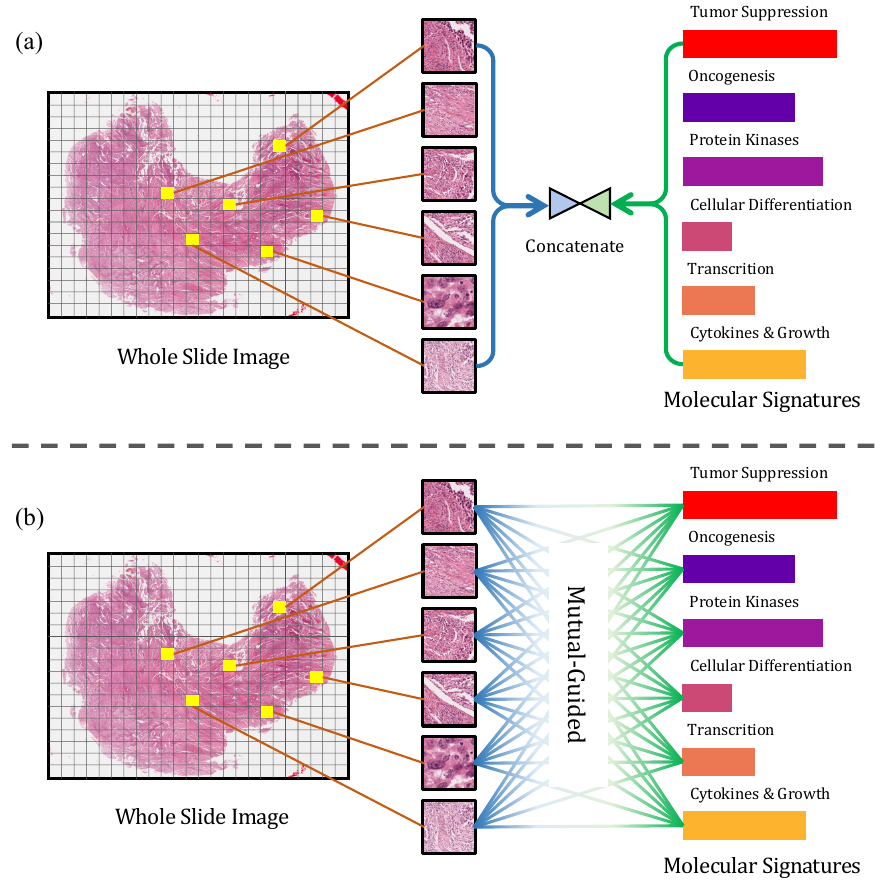}}
    \caption{Comparison between (a) Traditional early or late fusion and (b) Mutual-Guided Cross-Modality Attention-based integration. In (a), the WSI feature is concatenated with genomics simply, while in (b), histology and molecular data guide each other mutually, thereby capturing multimodal interactions.}
    \label{guide}
\end{figure}

Gigapixel whole slide images play a pivotal role in clinical cancer diagnosis, prognosis, and predicting patient response to treatment~\cite{yu2016predicting,campanella2019clinical,wulczyn2021interpretable}. However, the extensive gigapixel resolutions of WSIs present a challenge that existing methods primarily addressed through the utilization of a weakly-supervised multiple instance learning (MIL) approach. MIL involves the random sampling of image patches from the WSI as independent instances, followed by the application of a global aggregation operator to derive a representation at the bag level. Notably, previous studies have demonstrated the efficacy of MIL-based approaches in tackling \emph{needle-in-a-haystack} problems, such as cancer grading and subtyping, by solely leveraging slide-level labels without requiring detailed knowledge~\cite{thandiackal2022differentiable,zheng2022kernel,bian2022multiple}.

Prognosticating cancer, however, poses a formidable challenge as it necessitates the consideration of not only instance-level but also slide-level features pertaining to the tumor and its surrounding environment in order to evaluate the patient's relative risk of mortality~\cite{chen2021whole}. Early studies on the prediction of survival outcomes in cancer have predominantly relied on either genomic data or whole slide images. 
For genomic-based methods, such as the procedure developed by Bair \emph{et al}.~\cite{bair2004semi}, employ both gene expression data and clinical information to diagnose.
Katzman \emph{et al}.~\cite{katzman2018deepsurv} introduce a Cox proportional hazards deep neural network to model interactions between a patient’s covariates and treatment effectiveness for cancer prognosis.
Meanwhile, Huang \emph{et al}.~\cite{huang2019salmon} proposed an algorithm that consolidates and simplifies gene expression data and cancer biomarkers to facilitate prognosis prediction.
Conversely, whole slide images offer intricate morphological details. However, current survival analysis methods based on WSIs remain largely restricted to MIL approaches~\cite{ilse2018attention,yao2020whole,chen2021multimodal,chen2022pan}. 
For example, Yao \emph{et al}.~\cite{yao2020whole} introduced a Siamese MIL-based network to detect phenotypes associated with patients' survival outcomes.
Sandarenu \emph{et al}.~\cite{sandarenu2022survival} developed a MIL-based approach solely utilizing WSIs to make survival predictions in breast cancer. 

\label{sec:method}
\begin{figure*}[t!]
    \centering
    \centerline{\includegraphics[width=1\textwidth,height=0.55\textwidth]{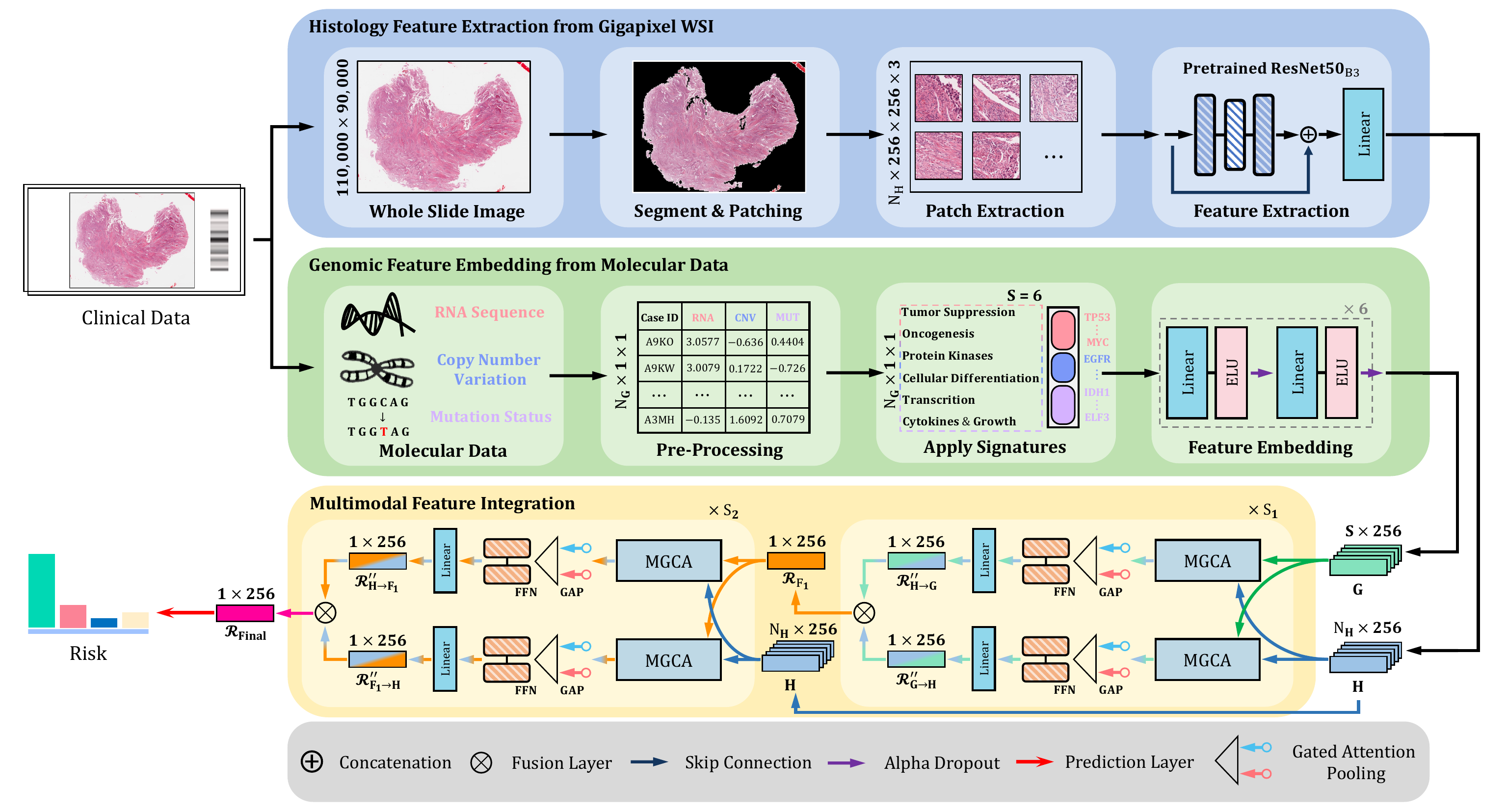}}
    \caption{Overview of the \textbf{M}utual-\textbf{G}uided \textbf{C}ross-Modality \textbf{T}ransformer (MGCT) framework. Initially, MGCT utilizes two independent streams – one for histology feature extraction and the other for genomic feature embedding – to tokenize the representations from diverse features, namely, gigapixel whole slide images and molecular data. Subsequently, the resulting multimodal feature embeddings from the two streams are integrated using two-stage MGCT layers to enable final survival outcome prediction. Note that $S_{1}$ and $S_{2}$ signify the number of MGCT layers employed in the fusion stage.}
    \label{MGCT}
\end{figure*}
In the current state-of-the-art, the gold standard for estimating patient survival relies on pathologists' manual assessment of histology and genomics~\cite{chen2021multimodal}. Consequently, there is a growing interest in multimodal learning methods that combine histopathology and genomic data for survival prediction, which have been gaining popularity~\cite{li2018graph,chen2021whole,chen2020pathomic,chen2021multimodal,chen2022pan}. However, these methods face significant challenges due to the large data heterogeneity gap between whole slide images and genomics. WSIs are typically represented as images with dimensions of 150,000$\times$150,000 pixels or as extensive bags containing tens of thousands of sampled patches as instances, while genomic features are usually represented as 1$\times$1 tabular data. As a result, existing early-fusion and late-fusion multimodal methods struggle to effectively learn the multimodal interaction features of these two highly disparate data types.

To address the above challenges, we propose a weakly-supervised, attention-based multimodal fusion framework (shown in Fig.~\ref{MGCT}), called \textbf{M}utual-\textbf{G}uided \textbf{C}ross-Modality \textbf{T}ransformer (MGCT) for survival outcome prediction using integrative WSI and genomic features. 

The \textbf{main contributions} of this paper are as follows: 
\begin{enumerate}
\item To overcome the limitations of current methods, we present a novel multimodal feature integration method named MGCT, which is designed to capture genotype-phenotype interactions in the tumor microenvironment.
\item We introduce a mutual-guided cross-modality attention (MGCA) as an attention-based feature integration strategy to effectively combine WSI and genomic features.
By leveraging the interplay between these two essential clinical features, the MGCA enables them to mutually guide each other during the integration process.
\item We conduct extensive experiments on five benchmark datasets from The Cancer Genome Atlas (TCGA), the results demonstrate that our method consistently outperforms current state-of-the-art methods.
\end{enumerate}

\section{Methodology}
\subsection{Problem Formulation}
In this survival analysis task, we denote the whole slide image as $\mathrm{X}$, the feature vector of genomic attributes with the WSI as $g$, the overall survival time (in months) as $t \in \mathbb{R}^{+}$, and the right uncensorship status (death observed) as $c \in \{0, 1\}$. Hence, we can represent the observations for all patient samples on the dataset as a quadruple $\left\{\mathrm{X}_{i}, g_{i}, t_{i}, c_{i}\right\}_{i=1}^{N}$, where $N$ indicates the number of WSIs in the dataset. The main objective is to develop and optimize the function $\mathcal{T}(\cdot)$ responsible for integrating the WSI bag $\mathrm{X}_{i}$ and the feature vector $g_{i}$ to estimate the hazard function:
\begin{equation}
    \begin{aligned}
    \hat{t_{i}} &= \mathcal{T} \left( \mathrm{X}_{i}, g_{i}\right) \\
    &= \phi \Bigg (\xi \bigg (\rho \Big (\left[f \left(x_{1}\right), f\left(x_{2}\right), \cdots, f\left(x_{N_{i}}\right) \right], g_{i}\Big)\bigg)\Bigg)
    \end{aligned}
\end{equation}
where 
$f \left (\cdot \right)$ is an instance-level encoder that processes features for each instance independently,
$\rho \left (\cdot \right)$ is the method for multimodal feature aggregation, 
$\xi \left (\cdot \right)$ is a \emph{permutation-invariant} instances aggregator that aggregates information and pools the extracted features to a single bag-level feature embedding, 
and $\phi \left (\cdot \right)$ is a bag-level classifier to make final predictions.

\subsection{WSI and Genomic Feature Construction}
\subsubsection{Histology Feature Extraction}
In this study, we adopt bag construction methods commonly used in conventional MIL approaches. To process each WSI $\mathrm{X}_{i}$, we employ the CLAM open-source repository~\cite{lu2021data} for automated tissue segmentation. Following segmentation, we extract 256$\times$256 image patches $\left\{x_{k}\right\}_{k=1}^{N_{i}}$ without spatial overlapping at the 20$\times$ equivalent pyramid level from all tissue regions identified. To create feature embeddings for the extracted patches, we utilize an ImageNet-pretrained ResNet-50~\cite{he2016deep} as a CNN encoder $f(\cdot)$ to convert each 256$\times$256 patch into a $d$-dimensional feature embedding $h_{k} \in \mathbb{R}^{d \times 1}$, truncated after the third residual block and adaptive average pooling layer. Considering the $N_{i}$ patches from the WSI $\mathrm{X}_{i}$, we assemble the extracted patch embeddings into a WSI bag representation $\mathcal{H}_{i} \in \mathbb{R}^{d \times N_{i}}$.

\subsubsection{Genomic Feature Embedding}
Genomic features such as transcript abundance (bulk RNA-Seq), gene mutation status, and copy number variation are typically represented as 1$\times$1 measurements, denoted as $g_{i} \in \mathbb{R}^{1\times1}$. This kind of data exhibits a high-dimensional low-sample size (HDLSS) nature, containing hundreds to thousands of features with relatively few training samples. Consequently, traditional feed-forward networks are prone to overfitting when processing such data.
To address this challenge, we leverage the Self-Normalizing Neural Network (SNN)~\cite{klambauer2017self} to formulate the genomic feature embedding. The SNN architecture utilized for the molecular feature input consists of two hidden layers, each comprising 256 neurons with exponential linear units (ELU) activation and Alpha Dropout applied to every layer.  By employing the SNN, we obtain genomic embeddings $\{ g_{i} \in \mathbb{R}^{d \times 1}\}_{s=1}^{S}$. Subsequently, we aggregate the genomic embeddings into a genomic bag representation and structure them based on S related biological functional impacts, yielding $\mathcal{G}_{i} \in \mathbb{R}^{d \times S}$, where $S$ represents the unique functional categories of the genomic data.

\subsection{Mutual-Guided Cross-Modality Transformer Layer}
Current approaches for histology-genomic fusion often rely on early fusion or late fusion-based strategies to address the significant data heterogeneity gap between gigapixel WSIs and genomic data such as concatenation~\cite{chen2021multimodal,li2023survival}, bilinear pooling~\cite{chen2022pan}, and Kronecker product~\cite{chen2020pathomic}. However, these fusion mechanisms have limitations in capturing interactions between genomic molecular data and WSIs effectively. In order to bridge the data heterogeneity gap between whole slide images and genomic features while capturing meaningful interactions between genomic-based phenotypes and the tumor microenvironment within gigapixel WSIs, we introduce the Mutual-Guided Cross-Modality Transformer layer (denoted as $\textbf{MGCT-Layer} \left( \cdot\right)$). Given the WSI bag representation $\mathcal{H}_{i}$ and genomic bag representation $\mathcal{G}_{i}$, we propose a Mutual-Guided Cross-Modality Attention (MGCA) to generate a genomic-guided feature embedding $\mathcal{R}_{ G \to H}$. MGCA is a variant of Multi-Head Self-Attention (MHSA) within vanilla transformer encoder layer~\cite{ilse2018attention}, for MGCA, $\mathrm{Q}$ is derived from one modality's features while $\mathrm{K}$, $\mathrm{V}$ are obtained from another modality features, the remaining architectures and calculation are analogous to the MHSA.
Subsequently, we utilize a gated attention pooling operation~\cite{ilse2018attention} to aggregate the feature embedding and direct it to a feed-forward network, resulting in the enhanced genomic-guided WSI embedding $\mathcal{R}_{ G \to H}^{\prime \prime}$. 
The process of the Mutual-Guided Cross-Modality Transformer layer is formulated as follows:
\begin{equation}
    \begin{aligned}
        & \mathbf{MGCA} \left( \mathcal{G}_{i}, \mathcal{H}_{i}, \mathcal{H}_{i} \right) = \mathbf{Softmax} \left( \frac{\mathrm{Q} \cdot \mathrm{K}^{\top}}{\sqrt{d_{k}}}\right) \\
        & = \mathbf{Softmax} \left( \frac{\textbf{W}_{q} \cdot \mathcal{G}_{i} \cdot \mathcal{H}_{i}^{\top} \cdot \textbf{W}_{k}^{\top}}{\sqrt{d_{k}}} \right) \textbf{W}_{v} \cdot \mathcal{H}_{i} \longrightarrow \mathcal{R}_{ G \to H} \\
        & \mathcal{R}_{ G \to H}^{\prime} = \mathbf{AttnPool} \left( \sum\limits_{i=1}^N \boldsymbol{\alpha}_{i} \right) \cdot \mathcal{R}_{G \to H} \quad where \\
        & \boldsymbol{\alpha}_{i} = 
        \frac{\mathrm{exp} \left\{ \textbf{W} \Big( \mathrm{tanh} \left( \textbf{V} \cdot \mathcal{R}_{i}^{\top}\right) \odot \mathrm{sigm} \left( \textbf{U} \cdot \mathcal{R}_{i}^{\top}\right) \Big)\right\}}{\sum\limits_{j=1}^N \mathrm{exp} \left\{ \textbf{W} \Big( \mathrm{tanh} \left( \textbf{V} \cdot \mathcal{R}_{j}^{\top}\right) \odot \mathrm{sigm} \left( \textbf{U} \cdot \mathcal{R}_{j}^{\top}\right) \Big)\right\}} \\
        & \mathcal{R}_{ G \to H}^{\prime \prime} = \xi \Big( \mathbf{MLP} \left( \mathcal{R}_{ G \to H}^{\prime} \right) \textbf{W}_{\mathbf{MLP}} \Big) \textbf{W}_{\xi} \\
    \end{aligned}
\end{equation}
where $\textbf{W}_{q}$, $\textbf{W}_{k}$, $\textbf{W}_{v}$, $\textbf{W}$, $\textbf{V}$, $\textbf{U}$, and $\textbf{W}_{\mathbf{MLP}}$ are trainable weight matrices, $\boldsymbol{\alpha}_{i}$ is the learnable scalar weight for gated attention-pooling operation $\mathbf{AttnPool} \left( \cdot \right)$, $\odot$ is the element-wise multiplication, $\mathbf{MLP} \left(\cdot \right)$ is a multi-layer perceptron with two linear layers, $\mathcal{R}_{ G \to H}^{\prime \prime}$ is the enhanced genomic-guided WSI feature embedding for genomic domain conditioned on histopathology domain. 
Similarly, we derive an improved genomic embedding guided by the enhanced WSI through an additional parallel MGCT layer. Subsequently, we proceed with a fusion stage where the above two embeddings are aggregated to produce a fused mutual-guided feature embedding, as illustrated in
Fig.~\ref{MGCT}. To facilitate deeper integration of WSI-genomic multimodal features, we stack two fusion stages in succession. Notably, the output of the first fusion stage serves as one of the inputs for the subsequent stage. The comprehensive process for multimodal feature integration is outlined in Algorithm~\ref{alg1}.

\renewcommand{\algorithmicrequire}{\textbf{Input:}}
\begin{algorithm}[ht!]
\caption{The MGCT Algorithm}
\label{alg1}
\hspace*{\algorithmicindent} 
\\
\textbf{Input:} \\
I. WSI bag representation $\mathcal{H}_{i} \in \mathbb{R}^{d \times N_{i}}$. \\
II. Genomic bag representation $\mathcal{G}_{i} \in \mathbb{R}^{d \times S}$. \\
IV. Number of the MGCT layers in two multimodal feature fusion stages, $S_{1}$ and $S_{2}$. 
\begin{algorithmic}[1]
\FOR{$s_{1}=1$ to $S_{1}$} 
\STATE $\mathcal{R}_{ G \to H}^{\prime \prime} \longleftarrow \textbf{MGCT-Layer} \left( \mathcal{G}_{i}, \mathcal{H}_{i}, \mathcal{H}_{i} \right)$ \\
$\mathcal{R}_{ H \to G}^{\prime \prime} \longleftarrow \textbf{MGCT-Layer} \left( \mathcal{H}_{i}, \mathcal{G}_{i}, \mathcal{G}_{i} \right)$
\ENDFOR
\STATE $\mathcal{R}_{F_{1}} \longleftarrow \mathbf{Concatenate} \left(\mathcal{R}_{ G \to H}^{\prime \prime},  \mathcal{R}_{ H \to G}^{\prime \prime}\right)$ 
\FOR{$s_{2}=1$ to $S_{2}$} 
\STATE $\mathcal{R}_{ F_{1} \to H}^{\prime \prime} \longleftarrow \textbf{MGCT-Layer} \left( \mathcal{R}_{F_{1}}, \mathcal{H}_{i}, \mathcal{H}_{i} \right)$ \\
$\mathcal{R}_{ H \to F_{1}}^{\prime \prime} \longleftarrow \textbf{MGCT-Layer} \left( \mathcal{H}_{i}, \mathcal{R}_{F_{1}}, \mathcal{R}_{F_{1}} \right)$
\ENDFOR
\STATE $\mathcal{R}_{\mathbf{Final}} \longleftarrow \mathbf{Concatenate} \left(\mathcal{R}_{ F_{1} \to H}^{\prime \prime},  \mathcal{R}_{ H \to F_{1}}^{\prime \prime}\right)$ 
\RETURN{final multimodal feature embedding $\mathcal{R}_{\mathbf{Final}}$}  
\end{algorithmic}
\end{algorithm}

\section{Experiments and Results}
\label{sec:setup}
\subsection{Datasets and Evaluation Metrics}
To validate the effectiveness of the proposed MGCT, we used five cancer datasets from The Cancer Genome Atlas (TCGA)\footnote{\url{https://portal.gdc.cancer.gov/}}, a public cancer data consortium that contains matched diagnostic WSIs and genomic data with labeled survival times and censorship statuses. In this work, we used the following five cancer types: Bladder Urothelial Carcinoma (BLCA, $N$ = 437), Breast Invasive Carcinoma (BRCA, $N$ = 1,023), LUAD (Lung Adenocarcinoma, $N$ = 516), GBMLGG (Glioblastoma Multiforme \& Brain Lower Grade Glioma, $N$ = 1,042), and UCEC (Uterine Corpus Endometrial Carcinoma, $N$ = 539). For each patient sample, we meticulously collected all diagnostic WSIs employed for primary diagnosis, resulting in a total of 3,557 WSIs (approx 5 TB of gigapixel images, 48 million patches). 

We meticulously paired molecular data, encompassing mutation status, copy number variation, and RNA-Seq abundance for each patient sample. To organize the gene features into gene embeddings, we relied on gene sets of gene families, which are categorized based on common features such as homology or biochemical activity from the Molecular Signatures Database~\cite{subramanian2005gene}. 
We employed six functional categories ($S$=6) to define the genomic embeddings: 1) Tumor Suppression, 2) Oncogenesis, 3) Protein Kinases, 4) Cellular Differentiation, 5) Transcription, and 6) Cytokines and Growth. 
\newcommand{\myvspace}{1pt}
\newcommand{\myboxsize}{0.94\textwidth}
\newcommand{\myarraystretch}{1.3}

\begin{table*}[h!]
	\caption{Concordance index (C-index) comparison with 17 cutting-edge survival analysis methods across 5 different cancer datasets. Averages $\pm$ standard deviations from five-fold cross-validation are reported. The best results and the second-best results are highlighted in \textbf{bold} and in \underline{underline}, respectively.
	}\label{cindex}
	\vspace{\myvspace}
	\renewcommand{\arraystretch}{\myarraystretch}
	\setlength\tabcolsep{6pt}
	\centering
	\resizebox{\myboxsize}{!}
	{\begin{tabular}{l|l|cccccc}
	\toprule[1pt]
	& Methods & BLCA & BRCA & LUAD & GBMLGG & UCEC & Overall \\
			\midrule 
            \multirow{4}{*}{\spheading{Genomic}} 
            & MLP 
            &  0.566 $\pm$ 0.047 &  \underline{0.588 $\pm$ 0.060} &  0.612 $\pm$ 0.042 &  0.806 $\pm$ 0.023 &  0.516 $\pm$ 0.065 &   0.618  \\
            & SNN~\cite{klambauer2017self}
            &  0.541 $\pm$ 0.016 &  0.466 $\pm$ 0.058 &  0.539 $\pm$ 0.069 &  0.598 $\pm$ 0.054 &  0.493 $\pm$ 0.096 &   0.527 \\ 
            & DeepSurv~\cite{katzman2018deepsurv}
            &  0.567 $\pm$ 0.049 &  0.598 $\pm$ 0.054 &  0.608 $\pm$ 0.026 &  0.810 $\pm$ 0.020 &  0.577 $\pm$ 0.058 &   0.632 \\
            & CoxRegression~\cite{kvamme2019time}
            & 0.591 $\pm$ 0.041 &  0.568 $\pm$ 0.077 &  0.574 $\pm$ 0.042 &  0.705 $\pm$ 0.014 &  0.464 $\pm$ 0.099 &   0.580 \\
            \midrule
            \multirow{5}{*}{\spheading{Pathology}} 
            & Deep Sets~\cite{zaheer2017deep} 
            &  0.500 $\pm$ 0.000 &  0.500 $\pm$ 0.000 &  0.496 $\pm$ 0.008 &  0.498 $\pm$ 0.014 &  0.500 $\pm$ 0.000 &   0.499 \\
            & Attention MIL~\cite{ilse2018attention}         
            &  0.536 $\pm$ 0.038 &  0.564 $\pm$ 0.050 &  0.559 $\pm$ 0.060 &  0.787 $\pm$ 0.028 &  0.625 $\pm$ 0.057 &   0.614 \\
            & CLAM~\cite{lu2021data} 
            & 0.565 $\pm$ 0.027 & 0.578 $\pm$ 0.032 & 0.582 $\pm$ 0.072 & 0.776 $\pm$ 0.034 & 0.609 $\pm$ 0.082 & 0.622 \\
            & DeepAttnMISL~\cite{yao2020whole}   
            &  0.504 $\pm$ 0.042 &  0.524 $\pm$ 0.043 &  0.548 $\pm$ 0.050 &  0.734 $\pm$ 0.029 &  0.597 $\pm$ 0.059 &   0.581 \\
            & Patch-GCN~\cite{chen2021whole} 
            &  0.560 $\pm$ 0.034 &  0.580 $\pm$ 0.025 &  0.585 $\pm$ 0.012 &  \underline{0.824 $\pm$ 0.024} &  \underline{0.629 $\pm$ 0.052} &   0.636 \\ 
            \midrule
            \multirow{10}{*}{\spheading{Multimodal}} 
            & Deep Sets (Concat)               
            &  0.604 $\pm$ 0.042 &  0.521 $\pm$ 0.079 &  0.616 $\pm$ 0.027 &  0.803 $\pm$ 0.046 &  0.598 $\pm$ 0.077 &   0.629 \\
            & Deep Sets (Bilinear)     
            &  0.589 $\pm$ 0.050 &  0.522 $\pm$ 0.029 &  0.558 $\pm$ 0.038 &  0.809 $\pm$ 0.027 &  0.593 $\pm$ 0.055 &   0.614 \\
            & Attention MIL (Concat)          
            &  0.605 $\pm$ 0.045 &  0.551 $\pm$ 0.077 &  0.563 $\pm$ 0.050 &  0.816 $\pm$ 0.011 &  0.614 $\pm$ 0.052 &   0.630 \\
            & Attention MIL (Bilinear) 
            &  0.567 $\pm$ 0.034 &  0.536 $\pm$ 0.074 &  0.578 $\pm$ 0.036 &  0.812 $\pm$ 0.005 &  0.562 $\pm$ 0.058 &   0.611 \\
            & DeepAttnMISL (Concat)           
            &  0.611 $\pm$ 0.049 &  0.545 $\pm$ 0.071 &  0.595 $\pm$ 0.061 &  0.805 $\pm$ 0.014 &  0.615 $\pm$ 0.020 &   0.634 \\
            & DeepAttnMISL (Bilinear)  
            &  0.575 $\pm$ 0.032 &  0.577 $\pm$ 0.063 &  0.551 $\pm$ 0.038 &  0.813 $\pm$ 0.022 &  0.586 $\pm$ 0.036 &   0.621 \\
            & PORPOISE~\cite{chen2022pan}    
            &  0.613 $\pm$ 0.021 &  0.563 $\pm$ 0.056 &  \textbf{0.621 $\pm$ 0.045} &  0.818 $\pm$ 0.011 &  0.622 $\pm$ 0.061 &   0.647 \\
            & MCAT~\cite{chen2021multimodal}             
            &  \underline{0.624 $\pm$ 0.034} &  0.580 $\pm$ 0.069 &  \underline{0.620 $\pm$ 0.032} &  0.817 $\pm$ 0.021 &  0.622 $\pm$ 0.019 &   \underline{0.653} \\
            \cmidrule{2-8}
            & \textbf{MGCT (Ours)} 
            & \textbf{0.640 $\pm$ 0.039} &  \textbf{0.608 $\pm$ 0.026} &  0.596 $\pm$ 0.078 & \textbf{0.827 $\pm$ 0.024} & \textbf{0.645 $\pm$ 0.039} &  \textbf{0.663} \\
			\toprule[1pt]
	\end{tabular}}
\end{table*}

To evaluate MGCT, we conducted training using 5-fold Monte Carlo cross-validation for each cancer type, in which each dataset was divided into 80/20 partitions for training and validation. The cross-validated concordance index (C-index) values across the validation splits were utilized to measure the predictive ability of the model in ranking the survival times of each patient.
Additionally, we utilized Kaplan-Meier curves to visually represent the quality of patient stratification and the log-rank test to determine the statistical significance of patient stratification.

\subsection{Implementation Details}
MGCT is implemented using PyTorch 1.13.1 and trained on a workstation equipped with an NVIDIA Quadro GV100 GPU for 20 epochs. During training, we follow the settings of~\cite{chen2021multimodal} to ensure a fair comparison. We employed the Adam optimization with a learning rate of 2e-4 and weight decay of 1e-5. Due to samples having varying bag sizes, we use a batch size of 1, with 32 gradient accumulation steps. We set $S_{1} = 1$ and $S_{2} = 2$ in two fusion stages for better performance. Our related code and corresponding models will be publicly made available at \url{https://github.com/lmxmercy/MGCT}.

\begin{figure}[t]
    \centering
    \centerline{\includegraphics[width=0.5\textwidth,height=0.4\textwidth]{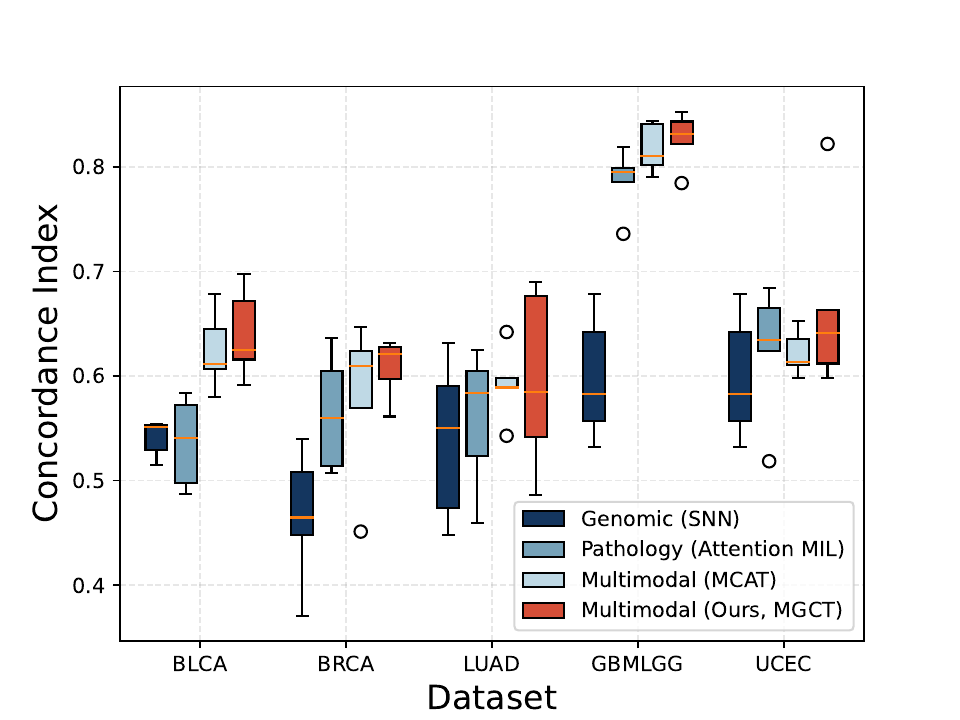}}
    \caption{C-index performance of SNN, Attention MIL, MCAT, and the proposed MGCT in 5 different cancer datasets in a 5-fold cross-validation.}
    \label{box}
\end{figure}

\subsection{Experimental Results}
We compare our method against the unimodal baselines and the multimodal SOTA methods are as follows:

\smallskip
\noindent{\bf Unimodal Baselines.}
For genomic data, we implemented an \textbf{MLP} as a traditional unimodal baseline, then we adapt \textbf{SNN}~\cite{klambauer2017self} which has been used previously for survival outcome prediction in the TCGA~\cite{chen2020pathomic,chen2021multimodal,xu2023multimodal}. Additionally, We utilized \textbf{DeepSurv}~\cite{katzman2018deepsurv} and \textbf{CoxRegression}~\cite{kvamme2019time} two genomic-only models for performance comparison. For pathology data, we compared the \textbf{Deep Sets}~\cite{zaheer2017deep} which is one of the pioneering neural network architectures for set-based deep learning, two SOTA MIL approaches \textbf{Attention MIL}~\cite{ilse2018attention}, \textbf{CLAM}~\cite{lu2021data}, and \textbf{DeepAttnMISL}~\cite{yao2020whole}, and a graph-based method for survival prediction \textbf{Patch-GCN}~\cite{chen2021whole}.

\smallskip
\noindent{\bf Multimodal Comparisons.}
As a multimodal comparison to the proposed \textbf{MGCT}, we enhanced the previous set-based
network architectures with concatenation and bilinear pooling, two common late fusion mechanisms to integrate bag-level WSI features and genomic features as multimodal baselines. We also compared \textbf{PORPOISE}~\cite{chen2022pan} and \textbf{MCAT}~\cite{chen2021multimodal} two SOTA methods for multimodal survival outcome prediction.

\smallskip
\noindent{\bf Unimodal versus Multimodal.}
Comparing the MGCT with all unimodal methods, MGCT achieved the highest performance in 4 out of 5 cancer datasets, indicating the effective integration of multimodal feature in our method. In comparison with the cutting-edge methods for genomic data, SNN and CoxRegression, MGCT outperformed them on all benchmarks, with overall C-index performance increases of 25.81\% and 14.31\%, respectively. Against the pathology baselines, MGCT improved on all the pathology-based unimodal approaches, with performance improvements in overall C-index ranging from 4.25\% to 32.87\% which demonstrating the merit of integrating both histopathology and genomic features. 
Although most multimodal methods were inferior to the unimodal genomic model in BRCA, GBMLGG, and UCEC datasets, the proposed MGCT achieved comparable performance to the genomic model in these cases.

\smallskip
\noindent{\bf Multimodal SOTA versus MGCT.}
MGCT outperformed on all multimodal approaches with an overall C-index performance increase ranging from 1.53\% to 8.51\%. In one-versus-all comparisons in each cancer dataset, MGCT achieved the highest C-index performance in 4 out of 5 cancer benchmarks, indicating its potential as a general method for any survival outcome prediction task. When compared with enhanced MIL-based multimodal methods using different fusion mechanisms, MGCT achieved a performance increase in overall C-index ranging from 4.57\% to 8.51\%, highlighting the effectiveness of the proposed multimodal feature integration method. Moreover, in comparison with the most similar work MCAT, in multimodal integration, MGCT demonstrated superior results on most cancer datasets, showcasing its ability to capture effective genotype-phenotype interactions in the tumor microenvironment, which are often crucial for cancer prognosis.

\begin{table*}[]
	\centering
	\caption{Quantitative results for ablation study on BLCA and UCEC two datasets. Deep Fusion: stack
two parallel MGCT layers in depth. MGCA: mutual-guided cross-modality attention. GAP: gated attention pooling operation in MGCT layer. Feedforward: position-wise feed-forward network in MGCT layer. We \textbf{bold} the highest performance.}\label{ablation}
	\vspace{\myvspace}
	\renewcommand{\arraystretch}{\myarraystretch}
	\setlength\tabcolsep{6pt}
	\resizebox{\myboxsize}{!}
	{\begin{tabular}{ccccccccc}
    \toprule[1pt]
    \multirow{2}{*}{Model} & \multicolumn{4}{c}{{Designs in MGCT}} & \multicolumn{2}{c}{TCGA-BLCA} 
    & \multicolumn{2}{c}{TCGA-UCEC}\\
    \cmidrule(lr){2-5}
    \cmidrule(lr){6-7}
    \cmidrule(lr){8-9}
    & Deep Fusion & MGCA & GAP & Feedforward & C-index $\uparrow$ & AUC $\uparrow$ & C-index $\uparrow$ & AUC $\uparrow$ \\ \midrule
    A & & & & & 0.499 $\pm$ 0.002 & 0.499 $\pm$ 0.002 & 0.499 $\pm$ 0.002 & 0.499 $\pm$ 0.002 \\
    B & \checkmark & & & & 0.535 $\pm$ 0.038 & 0.532 $\pm$ 0.045 & 0.541 $\pm$ 0.063 & 0.558 $\pm$ 0.034 \\
    C & \checkmark & \checkmark & & & 0.590 $\pm$ 0.045 & 0.598 $\pm$ 0.063 & 0.596 $\pm$ 0.037 & 0.615 $\pm$ 0.018\\
    D & \checkmark & \checkmark & \checkmark & & 0.601 $\pm$ 0.047 & 0.621 $\pm$ 0.072 & 0.608 $\pm$ 0.062 & 0.627 $\pm$ 0.071 \\
    E & \checkmark & \checkmark & \checkmark & \checkmark & \textbf{0.640 $\pm$ 0.039} & \textbf{0.679 $\pm$ 0.039} & \textbf{0.645 $\pm$ 0.039} & \textbf{0.660 $\pm$ 0.039} \\
	\toprule[1pt]
	\end{tabular}}
\end{table*}

\subsection{Patient Stratification}
\begin{figure*}[t!]
    \centering
\centerline{\includegraphics[width=1\textwidth,height=0.45\textwidth]{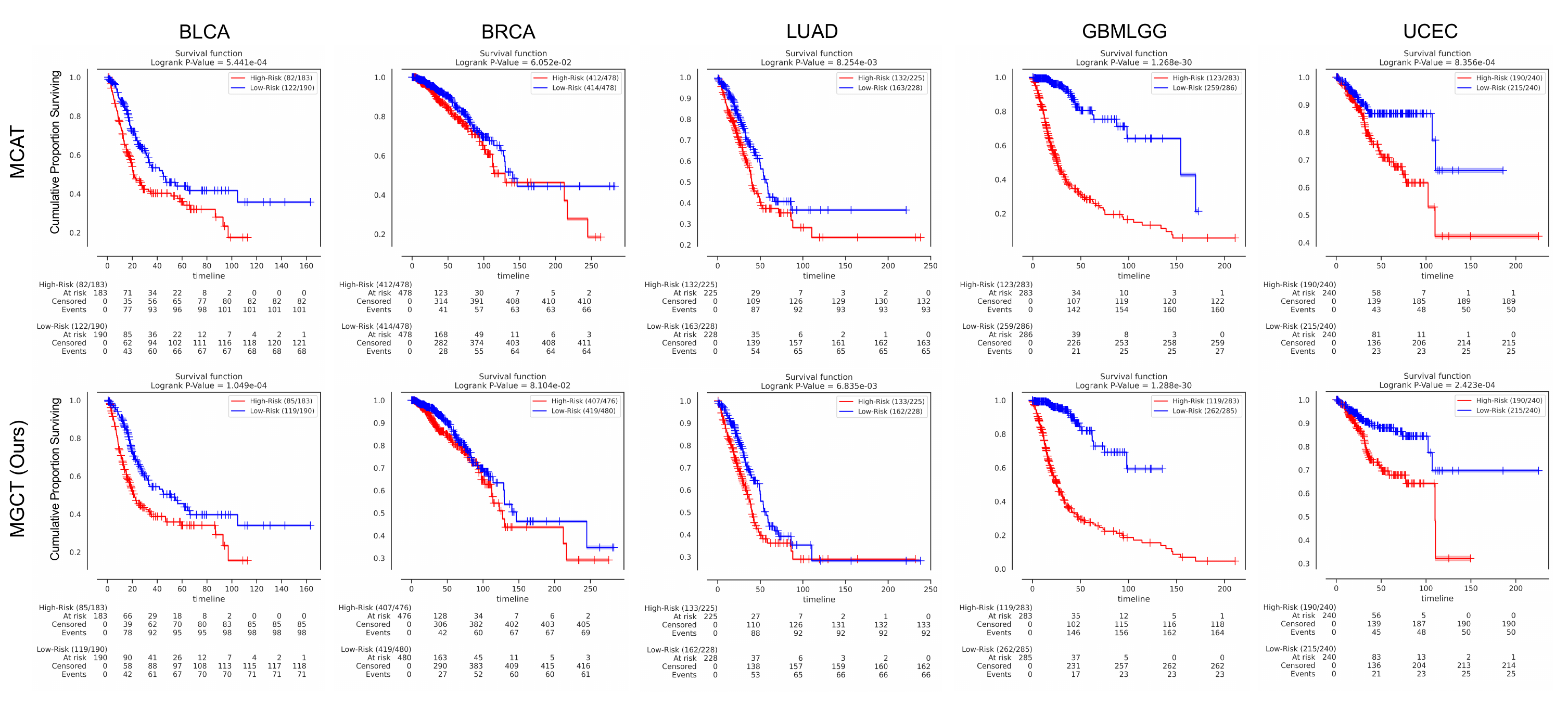}}
    \caption{Kaplan-Meier Analysis on five different cancer datasets, where patient stratifications of low risk (blue) and high risk (red) are presented. A P-value $<$ 0.05 indicates a significant statistical difference between the two groups, and a lower P-value is preferable. (Please zoom in for better viewing)}
    \label{km_curve}
\end{figure*}

To examine the patient stratification performance of MGCT, we generated Kaplan-Meier survival curves for different state-of-the-art methods. The patients were segregated into low-risk and high-risk groups based on predicted risk scores, and we presented the statistics on ground-truth survival time for each group in Fig.~\ref{km_curve}. The statistical significance test in the form of the Log-rank test was utilized to compare survival curves between low and high-risk patient groups, where a lower P-value indicated a more effective patient stratification. The results showed that when compared to competing methods, MGCT achieved clearer discrimination of the two risk groups for most datasets, demonstrating its superior performance in patient stratification.

\subsection{Ablation Study}
We conducted ablation studies on BLCA, GBMLGG, and UCEC three cancer datasets to validate the effectiveness of the proposed modules. We started with a basic model (Model A) based on the simple concatenation of WSI and genomic data.

\smallskip
\noindent{\bf Deep Fusion Strategy.}
We then investigated the effectiveness of the deep fusion strategy by creating Model B, which involved stacking two fusion stages in depth based on Model A. The C-index performance improvement of Model B over Model A was 7.21\% and 8.42\% respectively (Table~\ref{ablation}), which suggests that the deep fusion operation plays a critical role in enhancing the survival prediction performance, underscoring the importance of the proposed deep fusion strategy in MGCT.

\smallskip
\noindent{\bf MGCA Integration Strategy.}
To assess the effectiveness of mutual-guided cross-modality attention (MGCA) in the MGCT, we introduced Model C by incorporating the MGCA integration strategy into Model B. Notably, in TCGA-UCEC dataset, Model C exhibited a notable improvement of 10.17\% in C-index performance over Model B, This outcome underscores the significance of integrating MGCA, as it proves to be indispensable in enhancing the overall survival prediction performance of MGCT.

\smallskip
\noindent{\bf Gated Attention Pooling.}
We proceeded to augment Model C by incorporating the gated attention pooling (GAP) to create Model D, the experiment result highlighted the critical contribution of the gated attention pooling in MGCT. Removing the GAP operation (Model C) resulted in significant degradation of performance, notably affecting the C-index on two cancer datasets. This emphasizes the importance of the gated attention pooling in the overall effectiveness of MGCT.

\smallskip
\noindent{\bf Feed-forward Network.}
We also evaluate the effectiveness of the feed-forward network by adding it to Model D, then we compare the resulting Model E and Model D (w/o feed-forward network). The performance comparison proves that the network for improving the performance of survival prediction (6.49\% C-index and 9.34\% AUC improvement).



\section{Conclusion}
\label{sec:conclusion}
In this paper, we present an innovative weakly-supervised, attention-based multimodal learning framework called Mutual-Guided Cross-Modality Transformer (MGCT) for predicting survival outcomes in computational pathology. By leveraging mutual-guided cross-modality attention, we effectively integrate histology and genomic features to capture crucial genotype-phenotype interactions in the tumor microenvironment. The experimental results showcase the superiority of our proposed method over state-of-the-art approaches, underscoring its potential to significantly enhance survival outcome prediction in computational pathology.

{\small
\bibliographystyle{IEEEtranS}
\bibliography{egbib}
}



\end{document}